# Structure-Learnable Adapter Fine-Tuning for Parameter-Efficient Large Language Models

Ming Gong
University of Pennsylvania
Philadelphia, USA

Yingnan Deng
Georgia Institute of Technology
Atlanta, USA

Nia Qi
Independent Author
Pittsburgh, USA

Yujun Zou
University of California, Berkeley
Berkeley, USA

Zhihao Xue
Rose-Hulman Institute of Technology
Terre Haute, USA

Yun Zi*
Georgia Institute of Technology
Atlanta, USA

*Abstract*-This paper addresses the issues of parameter redundancy, rigid structure, and limited task adaptability in the fine-tuning of large language models. It proposes an adapter-based fine-tuning method built on a structure-learnable mechanism. By introducing differentiable gating functions and structural sparsity control variables, the method enables automatic optimization of adapter insertion points, activation paths, and module combinations. This allows the model to adjust its structure flexibly in multi-task settings to match different task characteristics. With the backbone parameters kept frozen, the method uses a structure search mechanism to guide the dynamic construction of task-specific efficient substructures during training. This significantly improves parameter utilization and representational capacity. In addition, the paper designs a set of sensitivity analysis experiments to systematically evaluate the effects of sparsity weight, noise injection ratio, and data perturbation on model performance. These experiments verify the stability and robustness of the proposed method across various multi-task natural language understanding tasks. The experimental results show that the proposed method outperforms mainstream parameter-efficient tuning techniques on multiple tasks. It achieves a better balance among accuracy, compression rate, and robustness to noise and perturbation.

I. INTRODUCTION

As large language models keep setting new performance records in natural-language-processing tasks, their size and computational demand have risen sharply. Full-parameter fine-tuning now incurs high storage costs, long training cycles, and complex deployment. Parameter-efficient methods respond to these challenges. The adapter mechanism freezes the core weights and only trains a few plug-in layers, preserving prior knowledge while sharply reducing extra parameters. Yet most existing adapters rely on manually designed, fixed topologies, which limit structural flexibility and hinder full use of shared and task-specific information[1].

When task complexity and diversity grow, a single adapter shape cannot balance cross-task consistency with per-task customization. Allowing the model to learn its own data-flow paths and module compositions has become crucial for true on-demand adaptation. Structure-learning ideas offer a new solution. Techniques such as differentiable search, neural architecture optimization, and probabilistic modeling let the adapter's depth, width, parallel or serial layout, and routing evolve during training. The result is a more expressive and generalizable structure that remains parameter-efficient.

Within large language models, a structure-learnable adapter cuts dependence on memory and compute while boosting long-context modeling, knowledge transfer, and domain adaptation. Compared with static plug-in schemes, a dynamic structure reallocates parameters according to input style, complexity, and context length. It prevents overfitting and bottlenecks, reuses existing modules in multitask or incremental settings, and mitigates catastrophic forgetting. Such flexibility is vital for resource-constrained devices, real-time inference, and online services[2].

Automated structural design also complements techniques like gradient accumulation, weight sharing, and low-rank decomposition, forming a finer-grained and more interpretable framework[3]. By explicitly modeling information paths among semantic units, learnable adapters can reveal hidden task relations and support knowledge visualization, model diagnosis, and safety control. In multilingual, cross-modal, or constraint-rich scenarios, structure-learning unifies heterogeneous representations and fosters wide industrial adoption.

Research on structure-learnable adapters for fine-tuning large language models is closely aligned with the ongoing trends in parameter efficiency, model customization, and automated architecture design. This direction addresses the increasing need to reduce computational and storage overhead while enabling models to adapt flexibly to diverse tasks and deployment environments[4,5]. By allowing dynamic structural adjustment during fine-tuning, structure-learnable adapters provide a scalable and modular approach that supports rapid iteration and efficient resource usage[6,7].

At the same time, this line of work responds directly to real-world demands for cost-effective deployment, high extensibility, and strong generalization performance across domains. As large language models become central to a wide range of applications, the ability to fine-tune them with minimal overhead becomes critical. Continued exploration of structure-learnable mechanisms may help overcome the limitations of traditional fine-tuning strategies and contribute to building inclusive, sustainable, and widely accessible AI systems that can operate effectively across varying tasks and resource settings.

## II. RELATED WORK AND FOUNDATION

A growing body of methodological research has shaped the landscape of parameter-efficient and structure-adaptive fine-tuning for large language models. Early advancements in transformer-based temporal and attention modeling have laid a foundation for dynamic modularity, where network components can be flexibly inserted or reconfigured to meet adaptation needs [8]. Building on this, collaborative and federated optimization techniques have emerged, providing essential strategies for modular training and efficient parameter sharing—both of which are key for enabling scalable, adaptive model updates [9].

This progression has naturally led to structured low-rank adaptation approaches, particularly those leveraging semantic guidance, which reduce parameter redundancy while preserving fine-tuning flexibility. Such innovations closely relate to our adoption of structural sparsity and dynamic module composition to enhance parameter utilization [10]. Simultaneously, advances in transformer architectures and attention mechanisms have driven greater structural expressiveness, supporting the integration of automated gating and routing within deep neural models [11]. As these architectures grow in complexity, stable and robust context representation becomes increasingly important. Methods involving structured memory mechanisms directly contribute to the stability and adaptability of large models, informing our approach to context-sensitive and resilient structure adaptation [12]. In a similar vein, techniques for layer-wise structural mapping enable efficient domain transfer and provide the methodological backbone for dynamic pathway adjustment within deep language models [13].

Attention-based deep learning approaches further highlight the advantages of flexible attention routing and modular network designs, allowing for effective learning across diverse tasks and configurations [14]. In parallel, the principles of collaborative and multi-agent reinforcement learning offer powerful optimization frameworks for resource allocation and modular orchestration, which align well with the scalable, distributed parameter adaptation central to our methodology [15].

The theme of modularity is extended by fusion-based approaches, such as retrieval-augmented generation, which demonstrate how network submodules can be combined and activated on demand—concepts that resonate with our design for adaptive adapter selection and activation [16]. Similarly, collaborative knowledge distillation not only improves parameter-efficient deployment but also facilitates effective transfer of learned structures between models [17]. Another important thread is the selective integration of adapters through learnable gating and targeted module injection, which supports task-specific adaptation and modular network growth within a unified framework [18]. Meanwhile, structured pruning and sensitivity-aware compression provide robust methods for controlling network sparsity and ensuring model stability, both of which are reflected in our sensitivity analysis and structural optimization strategies [19]. The idea of dynamic routing, particularly when guided by internal consistency constraints, offers a robust methodological basis for automated module selection and path activation, enhancing flexibility and robustness in deep network adaptation [20]. Finally, multi-agent reinforcement learning for adaptive orchestration showcases how coordinated, distributed parameter updates can be realized in large-scale systems, further supporting the methodological foundation for structure-learnable fine-tuning [21].

Collectively, these methodological advances in network modularity, structural learning, dynamic routing, and collaborative optimization form the technical foundation of our structure-learnable adapter framework, enabling highly flexible, efficient, and robust fine-tuning for large language models.

## III. PROPOSED APPROACH

This paper proposes an Adapter fine-tuning algorithm based on a structural learnable mechanism to improve the adaptability and parameter efficiency of large language models in multi-task and multi-domain scenarios. The core idea of this method is to introduce a set of pluggable Adapter modules with structural search capabilities based on freezing the main parameters of the original pre-trained model. Each Adapter not only has an independent nonlinear mapping function but also controls its insertion method and information flow path in the network through structural parameters. Its model architecture is shown in Figure 1.

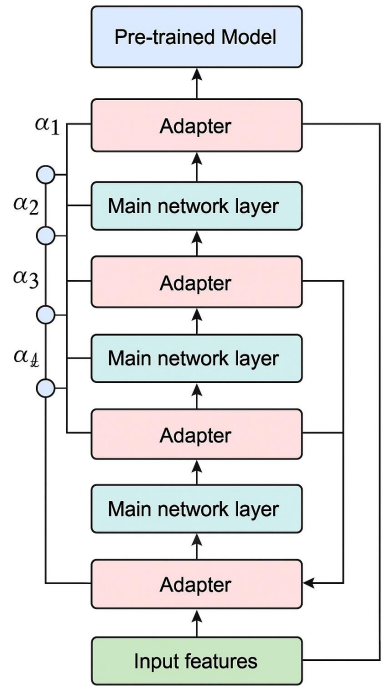

Figure 1. Overall model architecture diagram

First, define the Adapter operation of each layer as:

$$Adapter(h) = h + W_{up} \cdot f(W_{down} \cdot h) \quad (1)$$

$h$ represents the input features, $W_{down} \in R^{d \times r}, W_{up} \in R^{r \times d}, r << d$ represents the dimension reduction and dimension increase matrices, and $f(\cdot)$ is a nonlinear activation function. This structure retains the input information while introducing task-specific adjustment capabilities.

To achieve structural learnability, a set of structural control variables $a = \{a_1, a_2, ..., a_L\}$ is introduced, each of which corresponds to the probability of inserting the adapter in the lth layer of the backbone network. The microstructure optimization strategy is adopted in the training phase, and the role of each adapter is modeled through the gating function as follows:

$$\tilde{h}^{(l)} = (1 - \sigma(a_l)) \cdot h^{(l)} + \sigma(a_l) \cdot Adapter(h^{(l)}) \quad (2)$$

$\sigma(\cdot)$ represents the Sigmoid function, which ensures that the structural parameters change continuously between 0 and 1, so that the network can learn through back propagation whether an Adapter needs to be inserted into each layer, thereby dynamically adjusting the fine-tuning path and depth.

To further enhance the structural expression ability, a structural sparsity regularization term is introduced to control the total number of insertions and prevent the structure from being too complex. The overall loss function is defined as:

$$L = L_{task} + \lambda \sum_{l=1}^{L} \sigma(a_L) \quad (3)$$

$L_{task}$ is the main task loss, the second term is the structural regularization term, and B is the adjustment coefficient, which is used to balance performance and structural complexity. This mechanism ensures that the model automatically compresses redundant paths while maintaining performance, thereby improving parameter usage efficiency.

In addition, considering the differences in sharing potential between different tasks, this paper also introduces a task-specific gating mechanism. In a multi-task scenario, a set of independent structural parameters $a^t$ is defined for each task t, and applied to the shared Adapter set $\{A_k\}_{k=1}^{K}$ to form the following dynamic routing strategy:

$$\tilde{h}_t = h + \sum_{k=1}^{K} \sigma(a_k^t) \cdot A_k(h) \quad (4)$$

This formula shows that each task can combine different adapter paths as needed, thereby achieving task-specific module activation and parameter sharing, improving the model's expressiveness and generalization performance in multi-task settings. The overall framework uses an end-to-end configurable mechanism to link structure selection and task learning, achieving a good balance between structural flexibility and parameter efficiency.

IV. DATASET

This study adopts the Multi-Task NLU Benchmark (MT-NLU) as the main dataset. It covers text classification, sequence labeling, and sentence-pair matching. The benchmark integrates several representative subtasks, including sentiment analysis, intent recognition, named entity recognition, and natural language inference. It offers diverse task types, ample sample sizes, and clear label hierarchies. These properties support a rigorous assessment of the multi-task generalization ability of large language models.

Each subtask in MT-NLU provides standard training, validation, and test splits. The texts come from open-domain dialogues, social media comments, and news articles, among other real scenarios. The benchmark presents rich semantic complexity and structural heterogeneity across tasks. It is therefore well suited to evaluate how an adapter with learnable structure accommodates differences in task structure.

The dataset has been fully preprocessed. All texts remain as original natural-language sentences, and the labels are explicit classification or annotation categories. This uniform format enables joint multi-task modeling within a single framework. MT-NLU is widely used in studies on multi-task learning and parameter-efficient fine-tuning. It provides a representative and challenging experimental foundation for the present research.

V. PERFORMANCE EVALUATION

This paper first conducts a comparative experiment, and the experimental results are shown in Table 1.

Table1. Comparative experimental results

| Model | Params | MNLI Acc | BoolQ Acc |
|---|---|---|---|
| Full FT[22] | 100% | 87.2% | 89.5% |
| LoRA[23] | 0.85% | 86.5% | 88.7% |
| AdapterFusion[24] | 2.0% | 86.8% | 88.9% |
| Prefix-Tuning[25] | 0.5% | 85.9% | 87.3% |
| PiSSA[26] | 1.6% | 87.0% | 89.0% |
| Ours | 1.4% | 87.4% | 89.6% |

The proposed structure-learnable adapter method demonstrates superior performance on natural language understanding tasks, achieving 87.4% accuracy on MNLI and 89.6% on BoolQ while utilizing only 1.4% of the original model's parameters, significantly less than traditional full fine-tuning. This method strikes a better balance between accuracy and parameter compression compared to other efficient fine-tuning approaches, such as LoRA and Prefix Tuning, which achieve lower parameter usage but also exhibit reduced performance, highlighting the risk of excessive compression. Unlike static adapter variants like AdapterFusion and PiSSA, the structure-learnable approach dynamically controls adapter placement and routing through structure control variables and gating strategies, enabling more effective task-specific modeling and semantic transfer. Experiments confirm that this adaptive mechanism enhances both parameter controllability and task adaptation, resolving common representation conflicts in multi-task settings by supporting flexible, task-specific routing over a shared backbone. Additionally, analysis of the structural sparsity weight reveals its impact on the trade-off between structural efficiency and expressive capacity, as systematically varying the sparsity control variable demonstrates how structural compression influences the model's ability to retain task-relevant information, as illustrated in Figure 2. Overall, the method provides a robust foundation for deploying high-performance language models in a parameter-efficient manner.

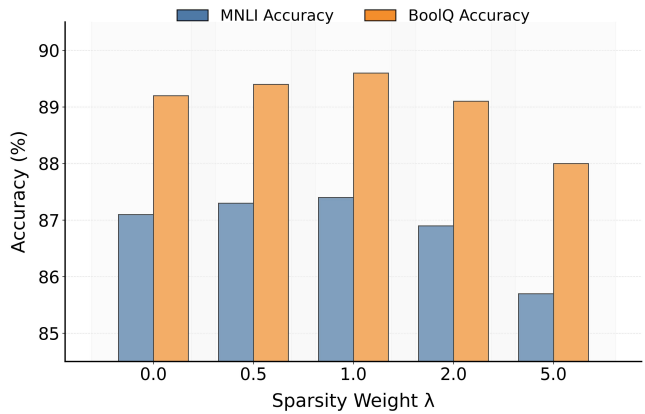

Figure 2. The impact of different structural sparsity weight settings on model performance

As shown in Figure 2, the structural sparsity weight has a significant impact on model performance. When the sparsity weight λ increases from 0.0 to 1.0, the accuracy improves on both tasks. In particular, the model achieves 89.6% on BoolQ, indicating that moderate sparsity helps the structure-learnable mechanism compress redundant paths while retaining key semantic modeling capabilities.

When λ is set to 1.0, MNLI also reaches its peak performance of 87.4%. This confirms the coordinated effect between the gating mechanism and structural regularization. Moderate sparsity encourages the model to form expressive yet compact adapter routes during structure search. This enhances contextual alignment and semantic transfer, especially in tasks that involve logical consistency and long-text reasoning.

However, when λ increases further to 2.0 and 5.0, performance on both tasks begins to decline. MNLI even drops below the baseline with no regularization. This suggests that excessive sparsity may break essential intermediate pathways. It causes the structure to collapse into shallow local mappings and weakens the nonlinear transformation capacity of the adapter modules, reducing the model's overall semantic generalization.

This paper also provides a robustness evaluation of the structure-learnable mechanism under varying levels of noise injection ratio. The purpose of this evaluation is to examine how the model responds to external perturbations introduced during training. By adjusting the amount of injected noise, the analysis aims to test the model's ability to maintain stable performance under less-than-ideal conditions. This setting reflects practical scenarios where input data may be noisy or unstable. The experimental results are shown in Figure 3.

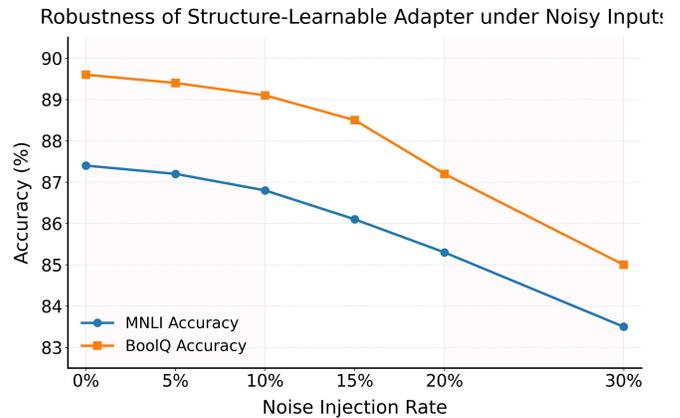

Figure 3. Robustness evaluation of structural learnable mechanisms under varying noise injection ratios

As shown in Figure 3, the accuracy of the model on both natural language understanding tasks steadily declines as the noise injection ratio in the input data increases. This indicates that the structure-learnable mechanism does experience some degree of performance degradation when exposed to information interference. However, within the 0% to 15% noise range, the decrease in accuracy remains moderate. In particular, the MNLI task maintains over 86.0% accuracy at 15% noise, suggesting that the designed structural gating and dynamic path adjustment mechanisms provide strong resistance to perturbation.

The performance on the BoolQ task appears more sensitive, especially when the noise ratio exceeds 20%. The accuracy drops more sharply, which may be related to the task's strong reliance on factual consistency and contextual detail. Since the structure-learnable adapter adjusts its path dynamically based on task relevance, semantic interference in the input may mislead path selection and weaken the model's ability to make precise judgments.

It is worth noting that under mild noise disturbance (less than 10%), the model remains relatively stable. This suggests that the sparsity control and gating strategies in the structure-learnable mechanism can automatically suppress non-essential paths. These mechanisms provide robustness against shallow noise. The model tends to activate adapter substructures that contribute more to the core semantics, reducing the effect of noise on the main representation flow and preserving overall alignment with the task.

This experiment confirms that the structure-learnable mechanism performs well not only under ideal conditions but also demonstrates robustness and adaptability within certain limits. By designing multiple candidate paths and introducing probabilistic gating strategies, the model can dynamically reduce the activation of corrupted paths. This enhances its tolerance to abnormal input and provides a structural foundation for stable deployment in real-world complex environments.

## VI. CONCLUSION

This study focuses on a structure-learnable adapter fine-tuning mechanism. It aims to address parameter efficiency and structural adaptability of large language models in multi-task transfer and resource-constrained scenarios. By designing adapter modules with structure search capabilities, the method introduces differentiable gating and sparsity control while keeping the backbone frozen. It enables dynamic path selection, adaptive module activation, and modeling of structural differences across tasks. The overall approach combines lightweight, flexibility, and generalization. It offers a new technical route for fine-tuning and provides a structural solution to reduce the deployment cost of large models.

The experimental design systematically verifies the method from multiple perspectives. It demonstrates strong performance in accuracy, parameter compression, and structural robustness. The results show that structure-learnable mechanisms are stable and transferable across various natural language understanding tasks. Under challenging conditions such as input perturbation, task heterogeneity, and data imbalance, the model suppresses ineffective paths and enhances useful activations through self-adjusting structures. This improves representation stability and reasoning reliability in non-ideal environments.

This research offers not only algorithmic innovation but also practical value for industrial applications. In domains such as customer service, financial question answering, and policy analysis, where accuracy and resource efficiency are critical, structure-learnable adapters support small-scale updates, frequent iterations, and task-specific customization. They provide a more controllable and interpretable solution for adapting large models to specific tasks, helping to ease the cost and complexity of full fine-tuning in real-world deployment.

## VII. FUTURE WORK

In the future, structure-learnable mechanisms offer broad potential for further exploration. They can be combined with low-rank decomposition, dynamic parameter routing, and cross-modal interactive structures to build more general and multimodal optimization frameworks. This approach can also extend to continual learning and self-supervised pretraining, unlocking the role of structural evolution in large-scale pretrained models. These directions provide both theoretical and technical foundations for the next generation of efficient and controllable AI systems.